\documentclass[10pt,twocolumn,letterpaper]{article}

\usepackage{cvpr}              % To produce the CAMERA-READY version
\usepackage{graphicx}
\usepackage{amsmath}
\usepackage{amssymb}
\usepackage{booktabs}
\usepackage{footmisc}  % Add this in the preamble
\usepackage[table]{xcolor}
\usepackage[pagebackref,breaklinks,colorlinks]{hyperref}
\usepackage[normalem]{ulem}  % Enables underlining
\hypersetup{
    colorlinks=true,
    urlcolor=blue
}

\usepackage{multicol} 
\usepackage[capitalize]{cleveref}
\crefname{section}{Sec.}{Secs.}
\Crefname{section}{Section}{Sections}
\Crefname{table}{Table}{Tables}
\crefname{table}{Tab.}{Tabs.}
\newcommand{\mysection}[1]{\vspace{2pt}\noindent\textbf{#1}}

\def\cvprPaperID{*****} % *** Enter the CVPR Paper ID here
\def\confName{CVPR}
\def\confYear{2025}

\global\long\def\comma{\,,}%

\begin{document}

\title{SoccerNet 2025 Challenges Results}
\author{
 \parbox{\linewidth}{\centering
 % Main lead
Silvio~Giancola$^{\dag1*}$,
Anthony~Cioppa$^{\dag2*}$,
 %
 % Challenge Organizers, alphabetic order
Marc~Guti\'errez-P\'erez$^{\dag3}$,
Jan~Held$^{\dag2}$,
Carlos~Hinojosa$^{\dag1}$,
Victor~Joos$^{\dag4}$,
Arnaud~Leduc$^{\dag2}$,
Floriane~Magera$^{\dag2,5}$,
Karen~Sanchez$^{\dag1}$,
Vladimir~Somers$^{\dag4,6,7}$
% ...
\\
\vspace{5mm} \normalsize
$^{1}$King Abdullah University of Science and Technology,
$^{2}$University of Li\'ege (ULi\'ege),
$^{3}$Institut de Robòtica i Informàtica Industrial (CSIC-UPC),
$^{4}$UCLouvain,
$^{5}$EVS Broadcast Equipment,
$^{6}$EPFL,
$^{7}$Sportradar,
% ...
}
}

\maketitle

%%%%%%%%% ABSTRACT
\begin{abstract}

\renewcommand{\thefootnote}{} % Remove numbering in the footnote
\footnotetext{$^*$Denotes equal contributions and $^\dag$ denotes challenges organizers.}
\renewcommand{\thefootnote}{\arabic{footnote}}  % Restore numbering after the footnote

The SoccerNet 2025 Challenges mark the fifth annual edition of the SoccerNet open benchmarking effort, dedicated to advancing computer vision research in football video understanding. This year’s challenges span four vision-based tasks: (1) Team Ball Action Spotting, focused on detecting ball-related actions in football broadcasts and assigning actions to teams; (2) Monocular Depth Estimation, targeting the recovery of scene geometry from single-camera broadcast clips through relative depth estimation for each pixel; (3) Multi-View Foul Recognition, requiring the analysis of multiple synchronized camera views to classify fouls and their severity; and (4) Game State Reconstruction, aimed at localizing and identifying all players from a broadcast video to reconstruct the game state on a 2D top-view of the field. Across all tasks, participants were provided with large-scale annotated datasets, unified evaluation protocols, and strong baselines as starting points. This report presents the results of each challenge, highlights the top-performing solutions, and provides insights into the progress made by the community. The SoccerNet Challenges continue to serve as a driving force for reproducible, open research at the intersection of computer vision, artificial intelligence, and sports.
Detailed information about the tasks, challenges, and leaderboards can be found at \href{https://www.soccer-net.org}{\uline{https://www.soccer-net.org}}, with baselines and development kits available at \href{https://github.com/SoccerNet}{\uline{https://github.com/SoccerNet}}.

\end{abstract}

\section{Introduction}
\label{sec:intro}

\begin{figure}[t]
    \centering
    \includegraphics[width=.49\linewidth,height=2.3cm]{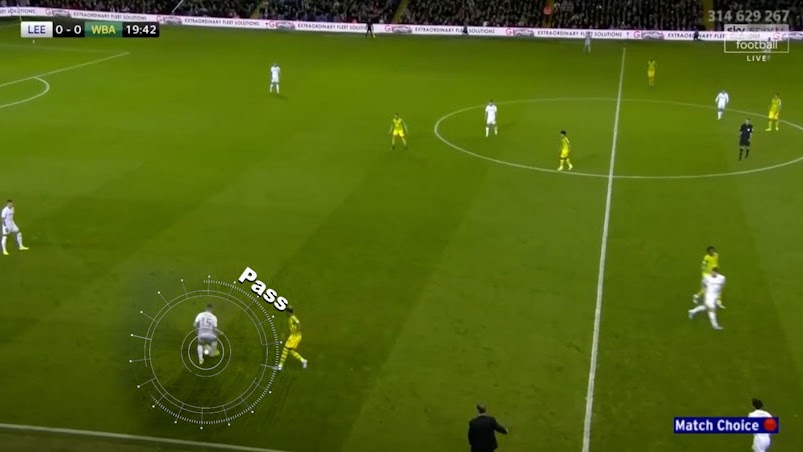}
    \includegraphics[width=.49\linewidth,height=2.3cm]{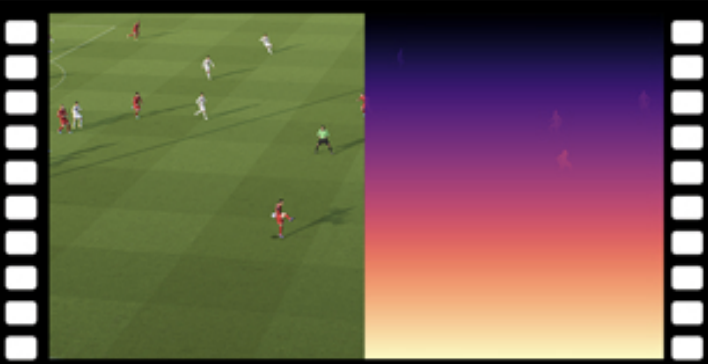}\\
    \includegraphics[width=.49\linewidth,height=2.3cm]{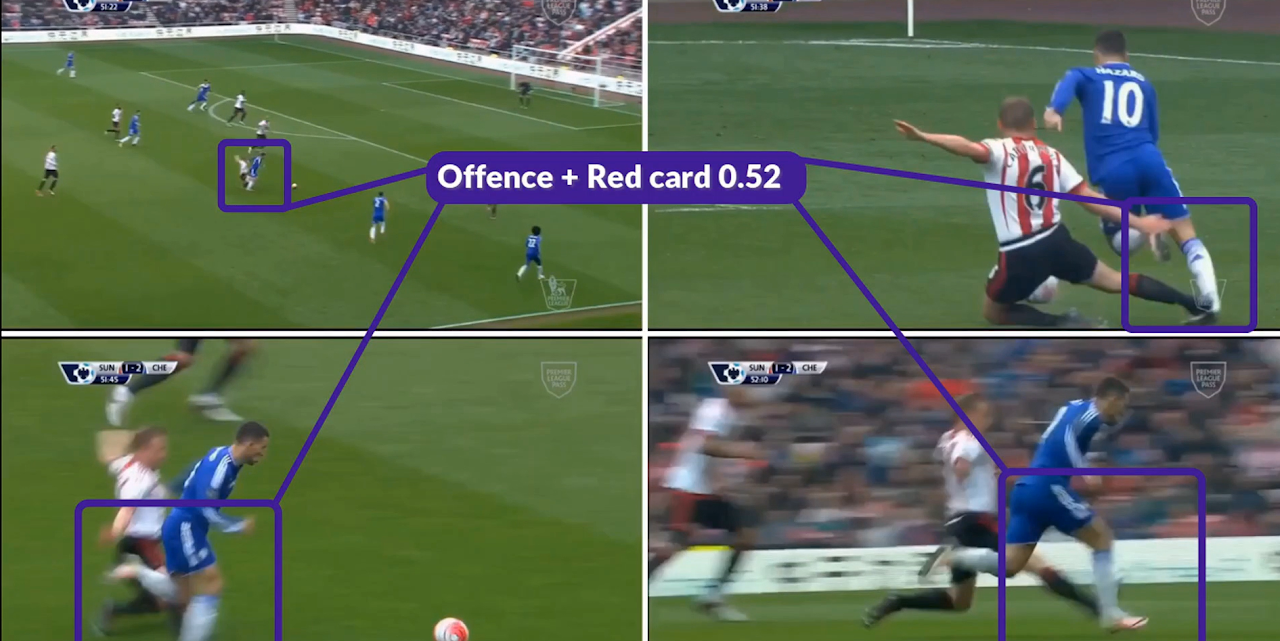}
    \includegraphics[width=.49\linewidth,height=2.3cm]{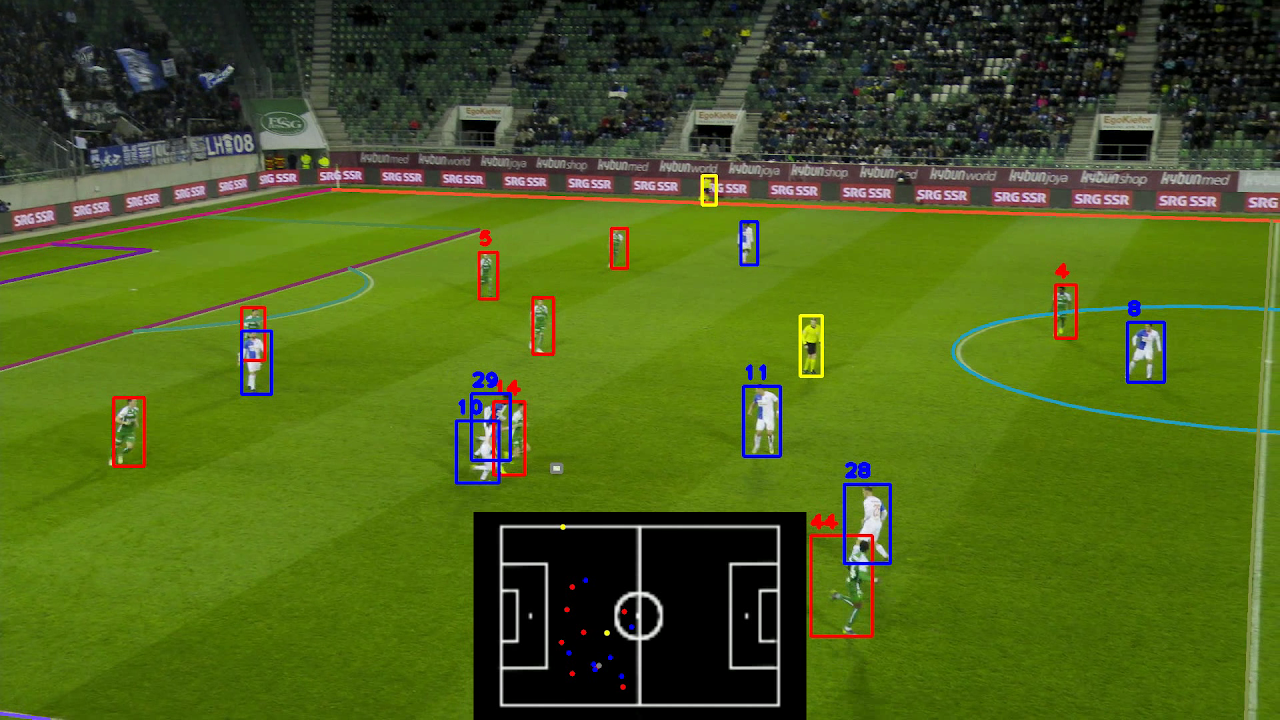}
    \caption{
In 2025, the challenges encompass four vision-based tasks.
(1) Team Ball Action Spotting, focusing on detecting ball-related actions in football broadcasts and assigning actions to teams,
(2) Monocular Depth Estimation, a novel task focusing on estimating relative depth maps from single-camera broadcast clips,
(3) Multi-View Foul Recognition, focusing on analyzing multiple synchronized camera views to classify foul incidents and assess their severity,
(4) Game State Reconstruction, focusing on reconstructing the game state, \ie, the player locations, roles, and team affiliations, onto a 2D top-view map of the field from broadcast video.
}
    \label{fig:graphical_abstract}
\end{figure}

The field of sports video understanding continues to evolve as a rich area of research in computer vision, with real-world applications spanning broadcasting, analytics, and decision support~\cite{Moeslund2014Computer, Gadde2024TheComputer}. At the center of this ecosystem, the SoccerNet benchmark has established itself as a leading resource, providing large-scale annotated datasets and a suite of open challenges that catalyze progress in automated football video understanding. Now in its fifth edition, the SoccerNet Challenges 2025 continue this tradition by bringing together academic and industry participants to tackle key problems in the understanding of football broadcasts.

This year’s challenges feature four vision-based tasks that cover complementary dimensions of the game: (1) Team Ball Action Spotting, focused on detecting ball-related actions in football broadcasts and assigning actions to teams; (2) Monocular Depth Estimation, targeting the estimation of relative depth maps from single-camera broadcast clips; (3) Multi-View Foul Recognition, requiring the classification of foul types and severities using multiple synchronized camera views; and (4) Game State Reconstruction, aimed at estimating the game state (player locations, identities, and roles) projected onto a 2D top-view minimap of the field.

These tasks span temporal, spatial, and semantic reasoning, each posing unique challenges in terms of volume of annotated data, model architectures, and training paradigms. Compared to previous years, the 2025 edition expands existing benchmarks with new data modalities, denser annotations, and more complex tasks.

This fifth edition of the SoccerNet Challenges saw vibrant participation across the board. The Team Ball Action Spotting competition attracted $79$ participants who submitted $67$ entries, with $16$ methods making it to the leaderboard. The Monocular Depth Estimation challenge saw $63$ participants and $111$ submissions, with $21$ final entries. The Multi-View Foul Recognition challenge had the highest engagement, with $88$ participants, $223$ submissions, and $24$ methods ranked. Finally, the Game State Reconstruction task drew $76$ participants and $66$ submissions, resulting in $14$ leaderboard entries. These numbers reflect the growing momentum of the SoccerNet initiative and the increasing interest in robust, reproducible benchmarks for sports video understanding.
\clearpage
\section{Team Ball Action Spotting}
\label{sec:ball_action_spotting}

\textit{Task managed and led by Artur Xarles.}

\subsection{Task description}

Team ball action spotting builds upon the ball action spotting task introduced in previous SoccerNet challenges~\cite{Cioppa2024SoccerNetChallenge, Cioppa2024SoccerNet2024Challenge-arxiv}. The objective is to temporally localize and identify all ball-related actions within untrimmed football videos. These include $12$ action classes: \textit{Pass}, \textit{Drive}, \textit{Header}, \textit{High Pass}, \textit{Out}, \textit{Cross}, \textit{Throw In}, \textit{Shot}, \textit{Ball Player Block}, \textit{Player Successful Tackle}, \textit{Free Kick}, and \textit{Goal}, where each action is annotated at a single temporal point corresponding to the exact moment the action occurs. Unlike previous editions, this year’s challenge introduces an additional layer of complexity: models must not only identify and localize actions, but also determine which team, on the left or right side of the field with respect to the camera view, is performing each action. This requires a deeper understanding of the game, including the ability to associate actions with specific players and to accurately identify their respective teams. The dataset includes $7$ annotated games from the English Football Leagues, with $2$ additional games reserved for challenge evaluation, for which annotations remain private.

\subsection{Metrics}

As in previous editions, the team ball action spotting task is evaluated using a metric based on the mean Average Precision (mAP)~\cite{Giancola2018SoccerNet}. In the standard formulation, the Average Precision (AP) is computed for each action class within a temporal tolerance of $\delta$, and the $mAP@\delta$ is obtained by averaging the AP scores across all classes. This year’s task introduces an additional challenge: identifying which team performs each action. Consequently, the evaluation metric has been adapted to account for team association. In the revised version, AP is computed separately for each combination of action and team. For each action class, a final AP is then calculated as a weighted average of the team-specific AP scores, with weights proportional to the number of ground-truth instances for each team. The overall $Team\text{-}mAP@\delta$ is finally obtained by averaging these action-level AP scores across all classes. To encourage precise temporal localization, the tolerance is set to $\delta = 1$ second.

\subsection{Leaderboard}

This year, $16$ teams participated in the team ball action spotting challenge, submitting a total of $67$ submissions. The best submission achieved a Team-mAP@1 of $60.03$, outperforming the proposed baseline, based on last year’s winning method, T-Deed~\cite{xarles2024t}, by $8.31$ points. The complete leaderboard can be found in Table~\ref{tab:performance_team_ball_action_spotting}.

\begin{table}[t]
    \centering
    \caption{
    \textbf{Team Ball Action Spotting Leaderboard.}
    The main evaluation metric, \textbf{$Team\text{-}mAP@1$}, is highlighted in bold. Best performances are also shown in bold. Teams with superscripts provided a summary, found in Appendix~\ref{app:TBAS} or Section~\ref{sub:ballactionspottingwinner} for the winning team.
    }
    \begin{tabular}{lcc}
        \toprule
        \bf Participant & \textbf{$\mathbf{Team\text{-}mAP@1}$} & $mAP@1$ \\
        \midrule
        dudek$^{TBAS-1}$ & \textbf{60.03} & 61.56 \\
        Intellindust-AI-Lab$^{TBAS-2}$ & 56.78 & 63.39 \\
        KIST-BAS & 56.08 & 58.09 \\
        kistimrcteam & 55.98 & 58.22 \\
        FSI-Tahakom$^{TBAS-5}$ & 55.51 & 63.89 \\
        liujian & 55.41 & 63.21 \\
        UniBW Munich VIS & 55.36 & \textbf{68.92} \\
        testken & 53.90 & 60.26 \\
        NYCU AINT Lab & 52.56 & 59.04 \\
        \rowcolor{gray!15} Baseline & 51.72 & 58.38 \\
        bupt miclab$^{TBAS-10}$ & 46.21 & 52.59 \\
        lunaandendymion & 39.74 & 47.71 \\
        deepvisionary & 11.67 & 25.48 \\
        poeclim & 11.67 & 25.48 \\
        GOLEM & 1.93 & 3.93 \\
        xxxyyye & 1.90 & 4.50 \\
        jennifer & 1.17 & 1.77 \\
        \bottomrule
    \end{tabular}
    \label{tab:performance_team_ball_action_spotting}
\end{table}

\subsection{Winner}
\label{sub:ballactionspottingwinner}

\mysection{TBAS-1}\\
\textit{Kamil Czarnogórski (int8@int8.io)}\\
The baseline T-DEED architecture was re-implemented within the open-source \texttt{dude.k} framework, enabling dataset access, clip generation, and reproducible training. Original dual heads for team classification and ball-action spotting were replaced by a unified head that predicts joint team-action classes, eliminating redundant non-action states and sharpening temporal boundaries. To expand training diversity, horizontal mirroring with label swapping, random crops, brightness jitter, and synthetic camera-pan transforms were employed. Training proceeded in two phases: large-scale pre-training on 500+ broadcast matches followed by dense fine-tuning on the challenge set. Bayesian hyperparameter optimization tuned learning-rate schedules, focal-loss weights, temporal anchor scales, and soft-NMS thresholds. The resulting configuration attained 60.03 Team-mAP@1 on the hidden challenge split.

\subsection{Results}\label{sec:ball_action_spotting_result}

Similar to the ball action spotting tasks from previous years, this year’s challenge continues to tackle the same main difficulties while introducing additional complexity. The task has been extended to not only detect the actions but also identify the team performing each one. As a result, it remains challenging to detect and precisely localize actions within the fast-paced nature of football, where only subtle visual cues indicate the brief moments in which actions occur. Additionally, the presence of $12$ different action categories, some with closely related semantics, further increases the difficulty of accurately distinguishing them. Moreover, the updated task now requires models not only to recognize which action is taking place, but also to determine which team is performing it.

The solutions presented by the participating teams for detecting and localizing actions generally followed similar trends to those introduced in last year’s edition of the challenge. Most approaches used a feature extractor to generate embeddings from video frames, combined with a temporal module to model long-range dependencies and capture actions across different temporal scales, which is important for recognizing actions with varying contexts. As in previous years, many methods also leveraged the provided SoccerNet Ball Action Spotting dataset, while additionally incorporating the larger original SoccerNet Action Spotting dataset, which contains 500 games with more sparsely annotated actions. This dataset was used either for pretraining or in joint training setups to improve performance. The main differences among this year’s methods stemmed from how they addressed the new component of the task: team identification. Some methods followed the baseline approach, using separate prediction heads to identify both the action and the associated team. Others adopted a unified prediction head that directly predicted combined action-team labels, treating each unique action-team pair as a distinct class. Alternatively, some approaches attempted to predict the team in possession at each frame, while others introduced mechanisms to enforce temporal consistency, encouraging sequences of actions typically performed by the same team to be assigned consistently. As shown in Table~\ref{tab:performance_team_ball_action_spotting}, the team prediction strategy was critical to the final ranking. Some teams performed well on the metric that ignored team identity, but they fell behind in the overall ranking, which was based on the team-aware evaluation metric. This highlights the importance of effectively modeling team information, as overlooking this aspect can heavily impact the performance of the final metric.

% \clearpage
\section{Monocular Depth Estimation}
\label{sec:moncoular_depth_estimation}

\textit{Task managed and led by Arnaud Leduc.}

\subsection{Task description}
Monocular Depth Estimation aims to predict a depth value for each pixel of every frame in team sports video sequences. These depth values represent the relative distance between objects in the scene and the camera, producing a depth map per frame. Unlike conventional datasets providing metric depth, the SoccerNet-Depth benchmark~\cite{Leduc2024SoccerNetDepth} relies on synthetic video frames rendered from two popular sports video games, where ground truth is available in relative scale only. As such, participants are required to estimate relative depth maps, invariant to scale, where depth values capture the ordinal structure of the scene rather than absolute distances. This year’s challenge focuses exclusively on monocular inputs, i.e., depth must be estimated from a single-view RGB camera without using multi-view information. Participants are evaluated on their ability to reconstruct dense relative depth maps that align with the ground truth, using standard depth metrics such as RMSE, AbsRel, and SILog. The dataset includes a public training set and a private challenge test set, covering diverse and dynamic scenes with occlusions, motion, and complex player-field interactions typical of broadcast sports footage.

\subsection{Metrics}
The Monocular Depth Estimation task is evaluated using multiple standard metrics widely adopted in the depth estimation literature. The metrics used are the Absolute Relative Error (Abs Rel), the Squared Relative Error (Sq Rel), the Root Mean Squared Error (RMSE), the Root Mean Squared Log Error (RMSE log), and the Scale-Invariant Log Error (SILog)~\cite{Eigen2014Depth}. Each metric is computed per frame and then averaged across the entire set. Among these, RMSE serves as the primary ranking criterion for the challenge. RMSE captures the average magnitude of error between predicted and ground-truth depth values by taking the square root of the mean of squared differences. It is sensitive to large deviations and provides an interpretable measure of overall prediction quality in pixel space. 
\begin{equation}
RMSE = \sqrt{\frac{1}{HW} \sum_{z_{ij} \in d} (z_{ij} - \hat{z}_{ij})^2}\comma
\label{eq:RMSE}
\end{equation}
where \textit{d} is a depth map of \textit{$H \times W$} pixels, \textit{$z_{ij}$} are the ground-truth depth values and \textit{$\hat{z}_{ij}$} are their corresponding predicted depth values.
To ensure consistency across submissions, participants are required to format their predictions as 16-bit PNG images, where depth values are first normalized to the range $[0, 1]$ and then scaled by $65,535$ before saving. Finally, predicted depth maps must follow the same orientation as the ground truth: pixels with smaller values represent points closer to the camera, while larger values correspond to more distant regions.

\subsection{Leaderboard}

In total, $21$ teams participated in the monocular depth estimation challenge, submitting a total of $111$ submissions. The best submission achieved a RMSE of $2.418\times10^{-3}$, outperforming the proposed baseline by $1.339\times10^{-3}$. The complete leaderboard can be found in Table~\ref{tab:performance_monocular_depth_estimation}.

\begin{table}[t]
    \centering
    \caption{
    \textbf{Monocular Depth Estimation Leaderboard.}
    The main evaluation metric, \textbf{$RMSE$}, is highlighted in bold. Best performances are also shown in bold. Teams with superscripts provided a summary, found in Appendix~\ref{app:MDE} or Section~\ref{sub:monoculardepthestimationwinner} for the winning team.
    }
    \resizebox{\linewidth}{!}{
    \begin{tabular}{lcc}
        \toprule
        \bf Participant & \textbf{$RMSE$}\scriptsize{$\times10^{-3}$} & $Abs Rel$\scriptsize{$\times10^{-3}$} \\
        \midrule
Hands-On Computer Vision$^{MDE-1}$ & \textbf{2.418} & \textbf{1.640} \\
HUST-Ipad$^{MDE-2}$ & 2.579 & 1.790 \\
bupt-miclab$^{MDE-3}$ & 2.683 & 1.860 \\
jacekm$^{MDE-4}$ & 2.746 & 2.070 \\
hvrl$^{MDE-5}$ & 2.816 & 2.280 \\
ucantseeme & 3.285 & 2.650 \\
dosensei & 3.285 & 2.650 \\
pengpai & 3.526 & 3.480 \\
\rowcolor{gray!15} Baseline & 3.757 & 4.020 \\
tradisoccernet & 3.757 & 4.020 \\
colanbw & 3.757 & 4.020 \\
liyiying & 3.757 & 4.020 \\
wangzhiyu918 & 3.823 & 4.030 \\
lipton & 3.849 & 3.270 \\
deleted\_user\_21545 & 5.480 & 5.250 \\
player & 9.487 & 12.880 \\
hihello & 9.487 & 12.880 \\
riq95560 & 12.666 & 15.100 \\
thelogical & 15.625 & 19.270 \\
dlogical & 16.919 & 20.560 \\
mrfahad & 21.815 & 27.120 \\
deleted\_user\_21510 & 35.365 & 53.810 \\
        \bottomrule
    \end{tabular}
    }
    \label{tab:performance_monocular_depth_estimation}
\end{table}

\subsection{Winner}
\label{sub:monoculardepthestimationwinner}

\mysection{MDE-1}\\
\textit{Fabian Perez, Henry Mantilla, Jorge García, Cristian Rey, Jose Sarmiento and Hoover Rueda-Chacón (\{perez2258059, henry2190071, jorge2180115, christhian2200190, jose2192232\}@correo.uis.edu.co, hfarueda@uis.edu.co)} 
Our solution introduces an effective fine-tuning approach leveraging the Depth Anything V2 pre-trained model~\cite{yang2024depthv2}. We fine-tuned the ViT-L architecture specifically for soccer scenes, optimizing it with a combination of Scale-and-Shift Invariant (SSI) loss and Scale-and-Shift Invariant Gradient Matching (SSIGM) loss \cite{li2025patchrefinerv2fastlightweight}. Training data, consisting of synthetic frames with $1918\times1078$ resolution, were preprocessed by converting relative depth $\mathbf{Y}$ to inverse relative depth $\bar{\mathbf{Y}}=1-\mathbf{Y}$ to match the architecture design. To tackle domain-specific challenges such as dynamic player pose and varied illumination, we incorporated tailored data augmentations, applying targeted color adjustments and geometric transformations. Optimized over seven epochs using AdamW, our final model demonstrated substantial improvements across all metrics compared to previous baselines. Code is avalaible at: \url{https://github.com/semilleroCV/Soccernet-depth-estimation-solution}.
\textbf{Acknowledgments.} We thank the financial support from the Faculty of Physical-Mechanical Engineering from the Universidad Industrial de Santander to attend CVPR.

 %[1] Yang, Lihe, et al. "Depth anything v2." Advances in Neural Information Processing Systems 37 (2024): 21875-21911.

 %[2] Li, Z., Cui, W., Bhat, S. F., & Wonka, P. (2025). PatchRefiner V2: Fast and Lightweight Real-Domain High-Resolution Metric Depth Estimation. arXiv preprint arXiv:2501.01121.

\subsection{Results}
The SoccerNet Monocular Depth Estimation Challenge 2025 invited participants to develop methods for predicting relative depth from single RGB frames of football scenes extracted from a popular video game. These methods were benchmarked against a prior baseline established using a fine-tuned ZoeDepth model, with substantial progress made across submissions. 
% MDE-1
Hands-On Computer Vision, led by Perez et al., introduced a fine-tuning pipeline based on DepthAnything V2 using full-resolution inputs (1918$\times$1078) and a combination of scale-and-shift invariant (SSI) and gradient matching (SSIGM) losses to preserve both global scale consistency and local depth discontinuities. Their domain-specific augmentations, including color transformations preserving field cues and horizontal flipping, yielded a model that surpassed the baseline across all metrics, 
% achieving an RMSE of 1.59×10$^{-3}$ and SILog of 0.261. 
achieving an RMSE of $2.418\times10^{-3}$. 
Their modifications confirmed that full-resolution inputs significantly improve predictions in dynamic, detail-rich scenes. 
% MDE-2
HUST-iPad developed MADE, a Motion-Aware Depth Estimation framework incorporating a motion-aware subtraction module and a weighted loss function to isolate dynamic foreground regions. Using consecutive frames, motion masks were computed and leveraged both for refinement via the Segment Anything Model and in a customized loss. A self-paced learning strategy further boosted performance by emphasizing difficult samples. MADE achieved an RMSE of $2.579\times10^{-3}$ and secured second place in the challenge. 
% MDE-3
BUPT MICLAB leveraged metric depth as a proxy for relative depth prediction by interpreting the football-relative depths as scaled metric values within a 0–256m range, considering that the depth distribution of frames is consistent, as they have similar viewpoints. Their approach involved freezing the DepthAnything, a zero-shot relative depth estimation model, encoder, and training a ZoeDepth-inspired decoder using a scale-invariant logarithmic loss. Only horizontal flipping was retained for data augmentation after an ablation revealed that color jittering and rotation degraded performance. Their best model achieved an RMSE of $2.683\times10^{-3}$ on the challenge set. 
% MDE-4
Jacek Maksymiuk proposed an innovative player-aware refinement pipeline focused on improving depth predictions, specifically in player regions. After fine-tuning DepthAnything on SoccerNet, player masks were extracted using YOLOv8-seg and passed to a secondary UNet with attention mechanisms. The residual between the DepthAnything prediction and ground truth was learned within these regions, yielding substantial local improvements. Maksymiuk’s final model (DepthAnything-player-ft) reached a 
% RMSE of 1.510$\times$10$^{-3}$ and SILog of 0.252, 
RMSE of $2.746\times10^{-3}$
% improving over the baseline by up to 57\% on certain metrics. 
% MDE-5
In a similar idea, team HVRL (Keio University) introduced a segmentation-guided refinement pipeline, integrating structural priors into the training. Using SAM2, they created masks for players (``person'') and the field (``grass''). These masks were used for region-specific loss weighting and post-processing, respectively. The player regions were emphasized during supervision, while field regions were smoothed using adaptive filters. Their method improved depth predictions notably in player zones and yielded strong quantitative results, 
% with an RMSE of 1.394$\times$10$^{-3}$ and SILog of 0.219 (Kageyama et al.). 
with an RMSE of $2.816\times10^{-3}$.
Together, these submissions showcase the importance of high-resolution training, motion awareness, targeted supervision, and structure-aware refinements in advancing the state of monocular depth estimation for complex, real-world sports scenes.

% \clearpage
\section{Multi-View Foul Recognition}
\label{sec:multi-view_foul_recognition}

\textit{Task managed and led by Jan Held.}

\subsection{Task description}
Multi-View Foul Recognition was introduced by Held~\etal~\cite{Held2023VARS} as a new computer vision task in sports and was first included in the 2024 SoccerNet challenge~\cite{Cioppa2024SoccerNet2024Challenge-arxiv}.
Since its introduction, automated foul classification has gained significant popularity in the sports community~\cite{Held2024Towards-arxiv, Held2024XVARS, Held2025Enhancing, Fang2024Foul-arxiv}.
The goal is to assist the decision-making process of a football referee by classifying whether an incident is a foul or not, the type of foul, and the severity of the foul. 
Therefore, participants must predict both the severity of the incident, \ie No offence, Offence + No card, Offence + Yellow card, and Offence + Red card, and its type, selecting from eight fine-grained foul categories among: Standing tackling, Tackling, High leg, Pushing, Holding, Elbowing, Challenge, and Dive/Simulation.
The task is structured as a multi-class classification problem: one for foul, one for severity assessment, and one for the foul type. It is built upon the SoccerNet-MVFouls dataset, which comprises $3,901$ labelled actions, each captured from two to four synchronised camera views at 224p and 720p resolution, recorded at $25$ frames per second.

\subsection{Metrics}
Given the class imbalance in the \textit{SoccerNet-MVFouls} dataset, we use the balanced accuracy as the primary evaluation metric. It is defined as:
\begin{equation}
    \text{Balanced Accuracy (BA)} = \frac{1}{N} \sum_{i=1}^{N} \frac{TP_i}{P_i}\comma
\end{equation}
where $N$ is the total number of classes, $ TP_i$ denotes the number of true positives for class $i$, and $P_i$ is the total number of ground-truth samples for that class.
To compute the final leaderboard score, we average the balanced accuracies of the two classification sub-tasks, foul type and severity, using:
\begin{equation}
    \text{combined\_metric} = \frac{BA_{\text{type}} + BA_{\text{off}}}{2}\comma
\end{equation}
where $BA_{\text{type}}$ measures performance on foul type classification, and $BA_{\text{off}}$ measures the performance of determining whether a foul occurred and its severity.

\subsection{Leaderboard}

This year, $24$ teams participated in the multi-view foul recognition challenge, with the winning team achieving a balanced accuracy of $52.22\%$, outperforming last year's winner by $7.46\%$.
The leaderboard can be found in
Table~\ref{tab:performance_mv_foul}.

\begin{table}[t]
    \centering
    \caption{
    \textbf{Multi-view foul recognition leaderboard}. 
    The main metric for the leaderboard and best performances
    are in bold. Teams with superscripts provided a summary, found in Appendix~\ref{app:MVF} or Section~\ref{sub:multiviewfoulwinner} for the winning team. Acc. stands for accuracy, BA. stands for balanced accuracy.
    $\dag$ indicates the winner of the previous year.
    }
    \resizebox{\linewidth}{!}{
    \begin{tabular}{l|lc|lc|c}
         & \multicolumn{2}{c|}{Type of Foul} & 
         \multicolumn{2}{c|}{Offence Severity} & \bf Combined \\ \midrule
    \bf Feat. extr.  & \bf Acc. & \bf BA. & \bf Acc. & \bf BA. & \bf BA. \\ \midrule
    UniBW Munich VIS$^{MVF-1}$ & $52.94$ & $\mathbf{48.38}$ & $38.91$ & $56.07$ & $\mathbf{52.22}$ \\
    zhangcc$^{MVF-2}$ & $53.39$ & $45.60$ & $39.82$ & $54.23$ & $49.91$ \\
    yeyee$^{MVF-3}$ & $61.54$ & $42.79$ & $47.06$ & $55.86$ & $49.32$ \\
    dwq123 & $43.89$ & $42.14$ & $41.63$ & $50.96$ & $46.55$ \\
    xiaoxiaoxiao & $36.20$ & $38.04$ & $39.82$ & $54.23$ & $46.13$ \\
    qazwsx & $44.34$ & $41.91$ & $40.72$ & $50.34$ & $46.12$ \\
    xxxyyye & $52.49$ & $40.02$ & $44.34$ & $51.72$ & $45.87$ \\
    danoj & $42.53$ & $37.67$ & $42.99$ & $53.64$ & $45.66$ \\
    WJB \dag  & $48.87$ & $30.88$ & $46.15$ & $\mathbf{58.65}$ & $44.76$ \\
    yuiop & $43.44$ & $37.26$ & $45.25$ & $50.64$ & $43.95$ \\
    chongcc & $43.44$ & $37.26$ & $45.25$ & $50.64$ & $43.95$ \\
    hahahahaha & $41.63$ & $37.11$ & $37.56$ & $50.42$ & $43.76$ \\
    axtonlu & $27.60$ & $35.59$ & $40.72$ & $49.22$ & $42.41$ \\
    ljd123 & $22.17$ & $34.63$ & $28.05$ & $44.09$ & $39.36$ \\
    leeli & $24.43$ & $30.58$ & $36.20$ & $46.79$ & $38.69$ \\
    taldarom & $\mathbf{62.44}$ & $42.92$ & $\mathbf{52.94}$ & $34.34$ & $38.63$ \\
    \rowcolor{gray!15} Baseline~\cite{Held2024Towards-arxiv} & $51.82$ & $39.52$ & $47.51$ & $34.46$ & $36.99$ \\
    player & $51.58$ & $39.52$ & $47.51$ & $34.46$ & $36.99$ \\
    vitomeme & $51.13$ & $39.03$ & $45.25$ & $33.79$ & $36.41$ \\
    odproject & $56.11$ & $35.55$ & $50.23$ & $32.97$ & $34.26$ \\
    linjiadong & $49.32$ & $37.86$ & $38.01$ & $28.47$ & $33.17$ \\
    kzkzkz$^{MVF-21}$ & $53.39$ & $45.60$ & $25.79$ & $17.81$ & $31.70$ \\
    knsg16$^{MVF-22}$ & $51.58$ & $36.35$ & $47.96$ & $25.69$ & $31.02$ \\
    soccer\_mamba & $21.27$ & $27.77$ & $54.75$ & $34.11$ & $30.94$ \\
    cola\_nbw & $9.50$ & $12.35$ & $41.63$ & $18.68$ & $15.51$ \\
    xxyycc & $15.38$ & $11.47$ & $25.34$ & $17.64$ & $14.56$ \\
    \end{tabular}
    }
    \label{tab:performance_mv_foul}
\end{table}

\subsection{Winner}
\label{sub:multiviewfoulwinner}

\mysection{MVF-1}\\
\textit{Konrad Habel, Fabian Deuser, Oliver Rose, Norbert
Oswald (konrad.habel@unibw.de, fabian.deuser@unibw.de, oliver.rose@unibw.de, norbert.oswald@unibw.de)}\\
Habel et al. leverage a single TAdaFormer-L/14~\cite{Huang2023TemporallyAdaptive-arxiv} model that was pre-trained on the Kinetics 400 and 710 datasets~\cite{Kay2017TheKinetics-arxiv}. 16 frames are sampled with a stride of two, resulting in a context window of 32 frames in total. For the aggregation of multiple views, max pooling is used before the classification head.
% Konrad \etal use a single TAdaFormer-L/14 \cite{Huang2023TemporallyAdaptive-arxiv} model pre-trained on the Kinetics 400 and 710 datasets\cite{Kay2017TheKinetics-arxiv}. 
% They use 16 frames with a step size two, resulting in a context window of 32 frames. For the aggregation of multiple views, max-pooling before the classification head is used. 
They incorporate view-specific context by adding a learnable view embedding to the feature vectors. This embedding is applied before the max pooling step to distinguish whether the input is a live view (clip 0) or a replay view (clip 1–3). 
The backbone gets fine-tuned in stage one using the 720p videos with an input size of 280×490 pixels, selecting random two of the up to four views per foul. For stage two, they extract
for all data ten times the features of the transformer before the classification heads using random augmentations and retrain in a second step only the classification heads on the
extracted features. In stage two the model sees all available views, which can close the gap between training and inference, leading to a slight improvement in performance.
The single model trained on all given data (train + valid + test) achieves on the challenge set a combined metric score of 49.52\% for stage one and 52.22\% for stage two.

\subsection{Results}
The SoccerNet 2025 MV-Foul Challenge introduced a range of novel methodologies that significantly advanced the field of automated foul recognition from multi-view football videos. A major innovation was the use of video large language models in a zero-shot setting, where participants demonstrated that prompting with rule-based context, dataset priors, and in-context examples can enable high-quality foul classification without any fine-tuning. 
Several teams proposed advanced aggregation strategies for multi-view video understanding. Notably, some participants introduced adaptive hybrid pooling, combining max and mean pooling techniques to enhance robustness.
Other participants leveraged view-specific embeddings and full-model fine-tuning of large-scale temporally adaptive transformers, achieving state-of-the-art results with a single trained model. 
Additionally, efficient training strategies were explored, including distributed multi-GPU setups and the use of pre-trained teacher models adapted to the task’s multi-label structure.
These approaches enabled more scalable and resource-aware solutions.

% \clearpage
\section{Game State Reconstruction}
\label{sec:game_state_reconstruction}

\textit{Task managed and led by Marc Gutiérrez-Pérez, Victor Joos, Vladimir Somers, and Floriane Magera.}

\subsection{Task description}
SoccerNet Game State Reconstruction (GSR) is a computer vision task that involves localizing and identifying players from a single moving camera to generate a video game-like top-view minimap of the state of the game, without requiring any specialized hardware worn by the players. This task is highly relevant to the sports industry, and the SoccerNet-GSR dataset~\cite{Somers2024SoccerNetGameState} now serves as the first and only open-source benchmark for evaluating methods in this domain.

The information to be extracted from the broadcast includes the following:
\begin{enumerate}
     \item The 2D position of all athletes on the field
     \item Their role (player, goalkeeper, referee, or other)
     \item The jersey number for each player and goalkeeper:
     \item The team affiliation (\ie, left or right w.r.t. the camera viewpoint) for each players and goalkeepers
\end{enumerate}

\subsection{Metrics}
Game State Reconstruction (GSR) performance is evaluated using the GS-HOTA~\cite{Somers2024SoccerNetGameState} metric, a task-specific extension of the well-established HOTA metric \cite{luiten2021hota}, widely used in multi-object tracking. GS-HOTA is designed to capture the unique challenges of the GSR task by jointly assessing the performance of spatial localization and identity matching of players on the field.
Unlike the original HOTA metric, which relies on bounding box Intersection-over-Union (IoU) in the image plane, GS-HOTA introduces a novel similarity measure tailored for GSR. This measure accounts for real-world 2D positions of players and incorporates additional identity attributes, namely jersey number, role, and team affiliation.
The GS-HOTA similarity score between a prediction (P) and a ground truth instance (G) is defined as:
\begin{equation} \label{eq:sim_gs_hota}
Sim_{\text{GS-HOTA}}(P, G) = \text{LocSim}(P, G) \times \text{IdSim}(P, G)\comma
\end{equation}
where the localization similarity is computed as:
\begin{equation} \label{eq:locsim}
\text{LocSim}(P, G) = e^{\ln(0.05)\frac{|P - G|_2^2}{\tau^2}}\comma
\end{equation}
and the identity similarity is defined as:
\begin{equation} \label{eq:idsim}
\text{IdSim}(P, G) =
\begin{cases}
1 & \text{if all attributes match,} \\
0 & \text{otherwise.}
\end{cases}
\end{equation}

This formulation ensures that both accurate localization and correct identification are rewarded. For further details and implementation specifics, we refer the reader to the work of Somers~\etal~\cite{Somers2024SoccerNetGameState}.

\subsection{Leaderboard}
In this second edition, $14$ teams participated in the game state reconstruction challenge, with a total of $66$ submissions. While the provided baseline was already improved the previous year, $12$ teams successfully surpassed it, achieving notable gains in performance. Specifically, the GS-HOTA score improved upon the baseline value from $29.01$ to $63.90$, achieved by the top-performing team. A summary of the final leaderboard is presented in Table~\ref{tab:performance_gsr}.

\begin{table}[t]
     \centering
     \caption{
     \textbf{Game state reconstruction leaderboard.}
     The main metric for the leaderboard and best performances
     are in bold. Team names with a superscript have
 provided a summary that may be found in
 Appendix~\ref{app:GSR} or in Section~\ref{sub:gamestatereconstructionwinner} for the winning
 team.}
     \resizebox{\linewidth}{!}{% <------ Don't forget this %
     \begin{tabular}{l|c|c|c}
 Participant team & \textbf{GS-HOTA (↑)} & GS-DetA (↑) & GS-AssA (↑) \\
     \midrule
     $\text{KIST-GSR}^{\textit{GSR-1}}$ & \textbf{63.90} & \textbf{51.36}	& 79.55 \\
     Constructor.Tech & 63.81 & 49.52 & 82.23\\
     $\text{lianyou}^{\textit{GSR-3}}$ & 62.76	& 46.17 & \textbf{85.33}\\
     $\text{Playbox \& MIXI}^{\textit{GSR-4}}$ & 61.64 & 45.04 & 84.36 \\
     tyler\_durden & 57.08 & 41.52 & 78.48\\
     SJTU Multi-Modal Perception Group & 50.06 & 32.98 & 75.99\\
     $\text{eidos.ai}^{\textit{GSR-7}}$ & 46.24 & 31.06 & 68.85\\
     $\text{UTokyo Football Lab}^{\textit{GSR-8}}$ & 40.13 & 21.33 & 75.51\\
     $\text{hjkim}^{\textit{GSR-9}}$ & 36.90 & 17.56 & 77.58\\
     junkwang & 33.62 & 18.09 & 62.51\\
     lsmuqi & 32.62 & 17.03 & 62.50\\
     hayden97 & 32.59 & 16.96 & 62.61\\
     \rowcolor{gray!15} Baseline~\cite{Somers2024SoccerNetGameState} & 29.01 & 13.71 & 61.41\\
     $\text{Laboratory for Biosignal Processing}^{\textit{
     GSR-13
     }}$ & 28.45 & 13.50 & 59.99\\
     johnstreet & 14.26 & 9.67 & 21.10\\
     \bottomrule

     \end{tabular}
     }
     \label{tab:performance_gsr}
\end{table}

\subsection{Winner}
\label{sub:gamestatereconstructionwinner}

\mysection{GSR-1}\\
\textit{Yin May Oo, Yewon Hwang, Vanyi Chao, 
Amrulloh Robbani Muhammad, Jinwook Kim}
\textit{(jinwook.kim21@gmail.com)}\\
Our game state reconstruction framework addresses player tracking and identity reconstruction in single-camera football broadcasts by integrating modular components for localization and identity recognition under challenging conditions. Athletes are detected via YOLO-X, tracked with Deep-EIoU and OSNet-based ReID embeddings, and mapped to pitch coordinates using a multi-frame keypoint detection model. Athlete identity is formulated as a multimodal autoregressive generation task with LLAMA-3.2-Vision, estimating role, jersey number, and color via instruction-based prompts for flexible, open-set recognition. Initial tracklets are refined using a split-and-merge strategy: splits are determined based on identity-prediction consistency; merges are guided by ReID similarity under spatial-temporal and identity constraints. Team affiliation (left/right) is inferred by clustering jersey colors and comparing cluster-wise mean x-positions, assuming players tend to remain on their side. Final post-processing improves identity consistency and team-level coherence using majority voting within tracklets and filtering unreliable trajectories, resulting in a robust and structured game state reconstruction.

\subsection{Results}
Although the GS-HOTA score of the top-performing team increased only marginally compared to the first edition of the challenge, the overall performance across the leaderboard improved substantially, fostering many novel and interesting approaches. For the object detection component, the use of fine-tuned detectors has become standard practice, with widely adopted models including YOLO-X~\cite{Ge2021YOLOX-arxiv}, YOLOv8~\cite{varghese2024yolov8}, YOLOv11~\cite{khanam2024yolov11}, and RF-DETR~\cite{rf-detr}. In the tracking stage, tracking-by-detection remains the predominant paradigm, with algorithms such as Deep-EIoU~\cite{huang2024iterative} and BoT-SORT~\cite{aharon2022bot} frequently employed. Several teams also integrated higher-level components such as GTA links~\cite{sun2024gta}, or adopted two-stage tracking strategies to further enhance performance. For the camera calibration task, popular approaches involved landmark detection followed by posterior calibration or homography estimation, with pipelines such as NBJW~\cite{GutierrezPerez2024NoBells}, PnLCalib~\cite{gutierrez4998149pnlcalib}, and Broadtrack~\cite{magera2025broadtrack} commonly used. A novel development in this edition was the incorporation of optical flow techniques to improve temporal smoothing. Significant innovations were also observed in player feature extraction for re-identification, role classification, team affiliation, and jersey number recognition. Notably, the use of CLIP- and OSNet-based features for these tasks, including object tracking, emerged as a strong approach. Furthermore, Vision Language models such as LLaMA-Vision~\cite{grattafiori2024llama} and Qwen2 VL Instruct~\cite{wang2024qwen2} were applied successfully for jersey number recognition.

% \clearpage
\section{Conclusion}
\label{sec:conclusion}

The SoccerNet Challenges 2025 marked a significant milestone in the evolution of vision-based football understanding, pushing the boundaries of what is possible with modern AI systems in sports analytics. Across four diverse tasks: Team Ball Action Spotting, Monocular Depth Estimation, Multi-View Foul Recognition, and Game State Reconstruction, participants tackled complex spatial, temporal, and semantic reasoning problems on broadcast football footage.

This fifth edition introduced several key innovations: increased granularity in action spotting with team attribution, the use of synthetic data for depth estimation, the adoption of multi-view inputs for semantic foul classification, and a comprehensive reconstruction of the game state from monocular views. These challenges demanded architectural creativity and thoughtful integration of domain knowledge and spatio-temporal context.

With over $280$ participants and hundreds of submissions across all tasks, the 2025 edition underscored the growing community interest and maturity of this research domain. Submissions consistently outperformed our proposed baselines, and several novel techniques, such as the use of VLMs, motion-aware depth estimation, and structure-aware tracking, set promising new directions for future research.

By continuing to offer open datasets, standardized evaluation protocols, and strong baselines, the SoccerNet Challenges remain a central hub for reproducible research and benchmarking in sports video understanding. We look forward to extending this momentum in 2026 with new tasks, further bridging the gap between computer vision research and real-world applications in sports.

% \textbf{Acknowledgement.} 
% The research reported in this publication was supported by funding from King Abdullah University of Science and Technology (KAUST) - Center of Excellence for Generative AI, under award number 5940.
% A. Cioppa is funded by the F.R.S.-FNRS. 
% This work was partly supported by the King Abdullah University of Science and Technology (KAUST) Office of Sponsored Research through the Visual Computing Center (VCC) funding and the SDAIA-KAUST Center of Excellence in Data Science and Artificial Intelligence (SDAIA-KAUST AI).

\clearpage

%%%%%%%%% REFERENCES
{\small
\bibliographystyle{ieee_fullname}
\bibliography{bib/abbreviation-short,bib/action,bib/camera-calibration-sports,bib/dataset,bib/labo,bib/learning,bib/soccer,bib/soccernet-challenge, bib/MDE-1, bib/sports,bib/NEW-REFS-HERE}}

\clearpage

\section{Supplementary Material}
\label{sec:supplementary_material}

\subsection{Team Ball Action Spotting}
\label{app:TBAS}

\mysection{TBAS-1}\\
\textit{Kamil Czarnogórski (int8@int8.io)}\\
The baseline T-DEED architecture was re-implemented within the open-source \texttt{dude.k} framework, enabling dataset access, clip generation, and reproducible training. Original dual heads for team classification and ball-action spotting were replaced by a unified head that predicts joint team-action classes, eliminating redundant non-action states and sharpening temporal boundaries. To expand training diversity, horizontal mirroring with label swapping, random crops, brightness jitter, and synthetic camera-pan transforms were employed. Training proceeded in two phases: large-scale pre-training on 500+ broadcast matches followed by dense fine-tuning on the challenge set. Bayesian hyperparameter optimization tuned learning-rate schedules, focal-loss weights, temporal anchor scales, and soft-NMS thresholds. The resulting configuration attained 60.03 Team-mAP@1 on the hidden challenge split.

\mysection{TBAS-2}\\
\textit{Jian Liu, Xuanlong Yu, Xi Shen, Di Yang (liujian2143@outlook.com, yu.xuanlong1996@gmail.com, shenxiluc@gmail.com, di.yang@ustc.edu.cn)}\\
We achieved a final mAP of 56.28 on the public leaderboard and 56.78 on the challenge phase. Our strategy is based on two key components: Baseline boosting and team alignment. Baseline boosting: We enhance the official baseline model through a class-mapping strategy and dataset mixture training. We mapping SoccerNet (SN) categories to SoccerNet Ball Action Spotting (SNB) categories, thus merging the double head baseline model to a single head one for action prediction. Then, we conduct a series of ablation studies on training the model using different dataset mixing ratios. The optimal one on the validation set we find is the model trained using 85\% SN and 15\% SNB dataset mixture. Furthermore, horizontal flipping for action prediction and class-specific ensembles for the PLAYER SUCCESSFUL TACKLE category further boost performance. Team alignment: Through the nature of football actions, we found that misclassifications in team prediction for certain actions, such as SHOT and GOAL, could be corrected by referencing the team assignment of correlated actions. For each action requiring correction, we look back at the preceding 4 seconds of video and identify any prerequisite actions. We then compare the confidence scores: if a prerequisite action exists, we assign the team label of the action with the highest confidence (between the current action and its prerequisites) to the rest of the actions. In the ablation study, we observe that this post-processing can greatly improve the mAP by considering different natural action series. In conclusion, by combining optimized dataset mixture, flipping, class-specific ensembles and team alignment, our approach significantly outperforms the baseline without modifying the model architecture.

\mysection{TBAS-5}\\
\textit{Faisal Altawijri, Ismail Mathkour, Saad Alotaibi (faltawijri@tahakom.com, imathkour@tahakom.com, sgalotaibi@tahakom.com)}\\
We propose a dual-backbone that utilizes T-DEED network for ball-action recognition that integrates two pre-trained models: Backbone A, trained on 12 action classes without team labels, and Backbone B, trained on the same video set with added binary team labels. Both backbones process the same 100-frame input clip, and their output features are concatenated. This fused representation feeds into three new heads: a 12-class for actions, a binary team classifier, and a displacement regressor for temporal refinement. All backbone weights remain frozen during training. To assess the impact of added capacity after fusion, four architectural variants were tested, ranging from identity mapping to a two-layer MLP. Each variant underwent two fine-tuning regimes: task-only training (4 videos with team labels) and joint training (including 500 action-only videos). A post-processing ensemble selected the best-performing model per class, improving team-mAP by ~3 points. Final results: mAP 63.89, team-mAP 55.51, with notable gains on rare actions like Penalty.

\mysection{TBAS-10}\\
\textit{Hongwei Ji, Hao Ye, Mengshi Qi, Liang Liu and Huadong Ma (\{hongweiji, haoye, qms, liangliu, mhd\}@bupt.edu.cn)}\\
To address the Ball Action Spotting task, we propose a multi-component framework to improve localization accuracy and recognition performance. Specifically, we first employ a feature extractor that encodes spatio-temporal representations from untrimmed video sequences. To better capture the temporal continuity around ball-related actions, we introduce a temporal refinement module that enhances temporal coherence and improves boundary precision. Finally, we incorporate a classifier augmented with a memory bank to perform dynamic score calibration and enable context-aware action prediction.

% \clearpage
\subsection{Monocular Depth Estimation}
\label{app:MDE}

\mysection{MDE-1}\\
\textit{Fabian Perez, Henry Mantilla, Jorge García, Cristian Rey, Jose Sarmiento and Hoover Rueda-Chacón (\{perez2258059, henry2190071, jorge2180115, christhian2200190, jose2192232\}@correo.uis.edu.co, hfarueda@uis.edu.co)} 
Our solution introduces an effective fine-tuning approach leveraging the Depth Anything V2 pre-trained model~\cite{yang2024depthv2}. We fine-tuned the ViT-L architecture specifically for soccer scenes, optimizing it with a combination of Scale-and-Shift Invariant (SSI) loss and Scale-and-Shift Invariant Gradient Matching (SSIGM) loss \cite{li2025patchrefinerv2fastlightweight}. Training data, consisting of synthetic frames with $1918\times1078$ resolution, were preprocessed by converting relative depth $\mathbf{Y}$ to inverse relative depth $\bar{\mathbf{Y}}=1-\mathbf{Y}$ to match the architecture design. To tackle domain-specific challenges such as dynamic player pose and varied illumination, we incorporated tailored data augmentations, applying targeted color adjustments and geometric transformations. Optimized over seven epochs using AdamW, our final model demonstrated substantial improvements across all metrics compared to previous baselines. Code is avalaible at: \url{https://github.com/semilleroCV/Soccernet-depth-estimation-solution}.
\textbf{Acknowledgments.} We thank the financial support from the Faculty of Physical-Mechanical Engineering from the Universidad Industrial de Santander to attend CVPR.

 %[1] Yang, Lihe, et al. "Depth anything v2." Advances in Neural Information Processing Systems 37 (2024): 21875-21911.

 %[2] Li, Z., Cui, W., Bhat, S. F., & Wonka, P. (2025). PatchRefiner V2: Fast and Lightweight Real-Domain High-Resolution Metric Depth Estimation. arXiv preprint arXiv:2501.01121.

\mysection{MDE-2}\\
\textit{Mingyang Du, Xian Li, Dingkang Liang, Hongkai Lin, Xin Zhou, Tianrui Feng, Dingyuan Zhang and
Xiang Bai (\{mydu, xianli01, dkliang\}@hust.edu.cn)}\\
In sports scenes, pronounced foreground–background disparities and varying depth scales often result in ambiguous object boundaries and unstable depth predictions. To tackle these challenges, we propose a Motion-Aware Depth Estimation (MADE) framework built on Depth Anything V2 and tailored for dynamic sports environments. MADE leverages temporal motion cues by computing inter-frame differences and applying threshold-based filtering to identify regions likely to undergo motion. These candidate regions are further refined by the Segment Anything Model (SAM) to ensure spatial alignment with the current frame, forming motion-aware masks. These masks are incorporated into the training loss to guide the model’s attention toward dynamic content. By injecting motion perception into the optimization process, MADE significantly enhances depth estimation accuracy, particularly in complex, fast-changing scenes.

\mysection{MDE-3}\\
\textit{Xiaoyang Bi, Shuaikun Liu, Zhaohong Liu, Yuxin Yang, Zhe Zhao, Mengshi Qi, Liang Liu and Huadong Ma (\{bxy, liushuaikun, liuzhaoh, yangyuxin, florasion, qms, liangliu, mhd\}@bupt.edu.cn)}\\
This paper presents our solution to the 2025 SoccerNet Monocular Depth Estimation Competition, which focuses on predicting relative depth in football scenes with limited training data. The ordinal relationships between points (i.e., relative depth) have attracted significant attention and are generally regarded as a more relaxed requirement compared to metric depth, thereby permitting the use of metric depth as a proxy for relative depth during both evaluation and training. We leverage the Depth Anything model, a powerful zero-shot monocular depth estimator pretrained on large-scale datasets and fine-tune it for metric depth estimation in football scenarios. This method enables effective learning despite the scarcity of labeled samples and bridges the gap between relative and metric depth estimation. Our method achieves a competitive score of $2.68 \times 10^{-3}$ on the challenge set, demonstrating the effectiveness of leveraging pretrained models and using metric depth as a proxy learning task.

\mysection{MDE-4}\\
\textit{Jacek Maksymiuk (maksymiukjacek1@gmail.com)}\\
We present a player-aware monocular depth estimation (MDE) pipeline tailored for soccer broadcast frames, developed for the SoccerNet MDE Challenge 2025. Our approach begins by fine-tuning the DepthAnything model on soccer-specific data to enhance global depth predictions. To further improve accuracy in critical regions, we apply a secondary correction model focused on soccer players. Player regions are detected using YOLOv8 segmentation, and depth refinement is achieved via a UNet with residual blocks and attention mechanisms. This correction model regresses residual depth values to better localize players in 3D space. Our approach improves depth estimation accuracy by 35–57\% across standard metrics compared to the SoccerNet baseline, with the player-aware model achieving the best overall performance.

\mysection{MDE-5}\\
\textit{Yuto Kageyama, Riku Itsuji,  Rintaro Otsubo, Kanta Sawafuji, Takuya Kurihara,  Tomoki Abe,  Hideo Saito \{y-kage.shadow96, diosa.de.gato.359, rintarootsubo, sawafuji\_kanta, takuyakurihara040130, abe.tomoki.1106, hs\}@keio.jp}\\
This method enhances depth estimation in high-resolution soccer videos by addressing two key challenges: inaccurate depth predictions around players and inconsistencies in field surface estimation. Building on the Depth Anything V2 model, the authors introduce a segmentation-guided refinement strategy. They leverage Grounded SAM2 to generate semantic masks for players (using the prompt "person") and the soccer field (using the prompt "grass").
During training, increased loss weights are assigned to player regions to prioritize accurate human depth estimation. At inference time, field regions undergo post-processing with median and average filters, applied only to valid depth pixels, to reduce noise while preserving edge details.
Given the dataset's low temporal resolution and high spatial fidelity, the approach is designed for per-frame (single-frame) depth estimation. Experimental results demonstrate notable improvements in both quantitative accuracy and qualitative depth map clarity, particularly around player regions. Importantly, the method does not depend on temporal continuity, making it well-suited for complex and dynamic sports video scenarios.

% \input{Appendix/MDE-8}

% \clearpage
\subsection{Multi-View Foul Recognition}\label{app:MVF}

\mysection{MVF-1}\\
\textit{Konrad Habel, Fabian Deuser, Oliver Rose, Norbert
Oswald (konrad.habel@unibw.de, fabian.deuser@unibw.de, oliver.rose@unibw.de, norbert.oswald@unibw.de)}\\
Habel et al. leverage a single TAdaFormer-L/14~\cite{Huang2023TemporallyAdaptive-arxiv} model that was pre-trained on the Kinetics 400 and 710 datasets~\cite{Kay2017TheKinetics-arxiv}. 16 frames are sampled with a stride of two, resulting in a context window of 32 frames in total. For the aggregation of multiple views, max pooling is used before the classification head.
% Konrad \etal use a single TAdaFormer-L/14 \cite{Huang2023TemporallyAdaptive-arxiv} model pre-trained on the Kinetics 400 and 710 datasets\cite{Kay2017TheKinetics-arxiv}. 
% They use 16 frames with a step size two, resulting in a context window of 32 frames. For the aggregation of multiple views, max-pooling before the classification head is used. 
They incorporate view-specific context by adding a learnable view embedding to the feature vectors. This embedding is applied before the max pooling step to distinguish whether the input is a live view (clip 0) or a replay view (clip 1–3). 
The backbone gets fine-tuned in stage one using the 720p videos with an input size of 280×490 pixels, selecting random two of the up to four views per foul. For stage two, they extract
for all data ten times the features of the transformer before the classification heads using random augmentations and retrain in a second step only the classification heads on the
extracted features. In stage two the model sees all available views, which can close the gap between training and inference, leading to a slight improvement in performance.
The single model trained on all given data (train + valid + test) achieves on the challenge set a combined metric score of 49.52\% for stage one and 52.22\% for stage two.

\mysection{MVF-2}\\
\textit{Chao Zhang, Jiadong Lin, Kexing Zhang, Lingling Li, Licheng Jiao, Long Sun, Wenping Ma (24171214014@stu.xidian.edu.cn)}\\
In this work, we propose a Multi-View Feature Fusion Network (MVFN). Our architecture employs Vision Transformer Large (ViT-L) as the backbone network for video feature extraction, initialized with pre-trained weights from an unmasked teacher foundation model. To address the challenge of inconsistent view counts across video samples, which complicates the design of adaptive fusion architectures, we implement multi-view feature fusion through our adaptive hybrid pooling multi-view fusion method. Recognizing that foul behaviors typically concentrate in specific temporal segments rather than spanning entire videos, we adopt a balanced training strategy by uniformly sampling 16 representative frames per video for efficient feature extraction and classification. Our method achieved a challenge phase score of 49.91 in the SoccerNet 2025 Multiview Foul Recognition Challenge, demonstrating effective performance trade-offs between computational efficiency and recognition accuracy

\mysection{MVF-3}\\
\textit{Yongliang Wu, Xinyu Ye, Bozheng Li, Yangguang Ji, Yi Lu, Yuyang Sun, Xinting Hu, Licheng Tang, Jiawang Cao, Jay Wu, Wenbo Zhu and Xu Yang (\{yongliangwu, yysun, xuyang\_palm\}@seu.edu.cn, 804677588@qq.com, xhu@mpi-inf.mpg.de, \{lee.li, jacob.ji, axton.lu, gary.tang, gavin.cao, jay.wu, vito.zhu\}@opus.pro)}\\
Automated soccer officiating remains a challenging computer vision task. Traditional deep learning approaches typically rely on extensive, task-specific annotations and often struggle to generalize or deeply internalize the complexities of the game. In this work, we propose a fully prompt-based framework for foul classification, leveraging Video Large Language Models (Vid-LLMs) in a zero-shot setting. Our main contribution is a meticulously crafted composite prompt that (1) positions the Vid-LLM as an expert referee, (2) provides concise summaries of IFAB rules, (3) incorporates statistical priors from the dataset, (4) employs Chain-of-Thought prompting to elicit transparent, step-by-step reasoning, and (5) integrates carefully selected textual few-shot examples from the training set for in-context learning. Without any additional training on the challenge-specific dataset, our structured prompting strategy successfully harnesses the multimodal reasoning capabilities of Vid-LLMs, achieving a final score of 49.32 on the SoccerNet 2025 MV-Foul challenge track.

\mysection{MVF-21}\\
\textit{Yuuki Kanasugi (kanasugi@star.rcast.u-tokyoac.jp)}\\
This report proposes a retrieval-augmented framework for foul classification in soccer, combining visual feature similarity with large language model (LLM) reasoning. First, spatio-temporal embeddings are extracted using the MViT v2-s baseline model and stored in a FAISS vector database along with metadata such as contact, body part, and intent. During inference, the top-5 most similar plays are retrieved based on a test clip’s features. Representative keyframes and metadata are then passed to ChatGPT-4o, which outputs two labels: foul presence and severity. Although the approach did not outperform the baseline in accuracy, it highlighted the potential of using LLMs to leverage rich contextual cues and improve interpretability. The study further revealed the importance of metadata quality, retrieval consistency, and prompt design in shaping LLM output. This framework offers a promising step toward more human-aligned, explainable decision-making in sports analytics.

\mysection{MVF-22}\\
\textit{Steven Araujo, Henry O. Velesaca and Abel Reyes-Angulo (\{saraujo, hvelesac\}@espol.edu.ec, areyesan@mtu.edu)}\\
In this work, we propose a novel Mamba-based multi-task framework for multi-view foul recognition, tailored for the SoccerNet Challenge 2025. Our approach leverages the Mamba architecture’s efficient long-range dependency modeling to process synchronized multi-view video inputs, enabling robust foul detection and classification in soccer matches. By integrating spatial-temporal feature extraction with a multi-task learning strategy, our model simultaneously predicts foul occurrences, identifies foul types, and localizes key events across multiple camera angles. We employ a hybrid loss function to balance classification and localization objectives, enhancing performance on diverse foul scenarios. Extensive experiments on the SoccerNet dataset demonstrate our method’s superior accuracy and efficiency compared to traditional CNN and Transformer-based models. Our framework achieves competitive results, offering a scalable and real-time solution for automated foul recognition, advancing the application of computer vision in sports analytics.

% \clearpage
\subsection{Game State Reconstruction}
\label{app:GSR}

\mysection{GSR-1}\\
\textit{Yin May Oo, Yewon Hwang, Vanyi Chao, 
Amrulloh Robbani Muhammad, Jinwook Kim}
\textit{(jinwook.kim21@gmail.com)}\\
Our game state reconstruction framework addresses player tracking and identity reconstruction in single-camera football broadcasts by integrating modular components for localization and identity recognition under challenging conditions. Athletes are detected via YOLO-X, tracked with Deep-EIoU and OSNet-based ReID embeddings, and mapped to pitch coordinates using a multi-frame keypoint detection model. Athlete identity is formulated as a multimodal autoregressive generation task with LLAMA-3.2-Vision, estimating role, jersey number, and color via instruction-based prompts for flexible, open-set recognition. Initial tracklets are refined using a split-and-merge strategy: splits are determined based on identity-prediction consistency; merges are guided by ReID similarity under spatial-temporal and identity constraints. Team affiliation (left/right) is inferred by clustering jersey colors and comparing cluster-wise mean x-positions, assuming players tend to remain on their side. Final post-processing improves identity consistency and team-level coherence using majority voting within tracklets and filtering unreliable trajectories, resulting in a robust and structured game state reconstruction.

\mysection{GSR-3}\\
\textit{Youxing Lian and Tiancheng Wang
(jy02624985@gmail.com, tiancheng.wang91@gmail.com)}\\
Our submission for the SoccerNet Game State Reconstruction challenge implements a four-stage process. We detect persons using YOLOv12 and 27 field line categories (16 points each) via a custom RT-DETR model. For field calibration, radial distortion correction (Wang et al.) is combined with PnLCalib (Gutiérrez-Pérez \& Agudo); fixed camera parameters are simplified to [pan,tilt,roll,focal] and temporally smoothed. Our tracking, enhancing IOF-Tracker (Liu et al.), utilizes  Farneback optical flow to improve Kalman filter-based motion prediction, especially during abrupt movements, applied to foreground objects identified via IoU and score thresholds. Features are extracted by CLIP-ReIdent (Habel et al., 0.5 cosine threshold). Finally, post-processing employs a lightweight 6-layer Transformer neural network to aggregate tracklet features, K-means for team assignment, and a fine-tuned ViT-L14 CLIP model (Habel et al.) for jersey number recognition.

\mysection{GSR-4}\\
\textit{Atom Scott, Calvin Yeung, Haruto Nakayama, Ikuma Uchida, Masato Tonouchi, Pragyan Shrestha, Rio Watanabe, Shunzo Yamagishi, Takaya Hashiguchi, Tomohiro Suzuki, Yuta Kamikawa
(\{atom, ikuma.uchida, pragyan\}@play-box.ai, \{yeung.chikwong, suzuki.tomohiro\}@g.sp.m.is.nagoya-u.ac.jp, nkym20020422@gmail.com, \{masato.tonouchi, rio.watanabe, takaya.hashiguchi\}@mixi.co.jp, yamagishi.shunzo@image.iit.tsukuba.ac.jp, ytkmkw@gmail.com)}\\
Our approach builds on the TrackLab framework, updating several modules with state-of-the-art techniques. For object detection, we use RF-DETR, outperforming the YOLO series. Feature extraction used CLIP-ReID embeddings, feeding our tracking pipeline consisting of BoT-SORT and Global Tracklet Association (GTA). Jersey number recognition closely follows previous work; we filter embeddings via Gaussian outlier rejection, geometrically extract torso regions using ViTPose keypoints, and recognize numbers using PARSeq OCR. Camera calibration extends BroadTrack by propagating calibration parameters temporally via optical flow, initialized from multiple stable frames. To address visual ambiguity or occlusion, we enhance role classification heuristically, assigning goalkeeper roles based on spatial positioning (furthest player in penalty areas), followed by greedy hill-climbing optimization penalizing invalid configurations, such as multiple goalkeepers. Finally, linear interpolation and a Savitzky–Golay filter are applied to ensure smoothness and handle short tracking gaps. This pipeline achieved 4th place (61.64 GS-HOTA). Future work will leverage domain-specific knowledge to enhance overall robustness.

\mysection{GSR-7}\\
\textit{Marcelo Ortega and Federico Méndez (\{ortegatron, federico\}@eidos.ai)}\\
% Game State Reconstruction leveraging Visual Large Models - eidos.ai
We boost player-ID accuracy by combining a large foundational vision–language model (Qwen2-VL-Instruct) with function calling. Instead of classifying each detection in isolation, we feed the model a long sequence of detections arranged on a grid. Through chain-of-thought reasoning, the model can invoke a “zoom” function to inspect specific frame ranges before choosing a jersey number, iterating until confident. We used this method also for the referee role recognition, and modified the clustering algorithm to search for a better balance between the number of tracks on the players clusters.

\mysection{GSR-8}\\
\textit{Sadao Hirose, Kazuto Iwai, Ayaha Motomochi and Ririko Inoue (\{hirose.akka, kazutoiwai.jp, ayaha.mochi5, lyrico1202\}@gmail.com)}\\
We present an enhanced pipeline built upon the SoccerNet 2025 GSR baseline. Images in the dataset were deblurred using DeblurGAN-v2. We fine-tuned a YOLOv8x detector on the SoccerNet dataset for improved athlete detection. Detected regions were then segmented with DeepLabV3 and MobileNetV3-Large to isolate player silhouettes. Segmented images were used to extract a 10-dimensional vector encoding uniform colors and their corresponding areas. These vectors augmented inputs to downstream modules for re-identification, role classification, and team identification. We also introduced GTA-Link to refine tracklets post-baseline tracking. A Gaussian blur filter was applied prior to jersey number recognition. After passing through the pipeline, we applied linear interpolation to fill in gaps in detections and stabilize trajectories. Similarly to the original baseline, our method had difficulties mostly with jersey number recognition. Future work includes incorporating multimodal large language models for more accurate jersey number recognition.

\mysection{GSR-9}\\
\textit{Hyungjung Kim, Joohyung Oh and Jae-Young Sim (\{hjkim1113, joohyung0809, jysim\}@unist.ac.kr)}\\
We developed four modules to improve the performance of tracking and re-identification. First, we applied motion deblurring to reduce the blurring artifacts in images and improve the detection performance of the players and jersey numbers. Second, we used optical flow estimation to compensate for the camera movements and enhance the stability of tracking. Third, we fine-tuned the detector model on SoccerNet-GSR dataset to further improve the performance. Lastly, we used uniqueness property that no more than 24 individuals (22 players and 2 referees) have different identities for reliable identity matching. The proposed method achieves a GS-HOTA score of 36.90\% in the challenge.

\mysection{GSR-13}\\
\textit{Julian Ziegler, Patrick Frenzel, Daniel Matthes and Mirco Fuchs (\{julian.ziegler, patrick.frenzel, daniel.matthes,mirco.fuchs\}@htwk-leipzig.de)}\\
This report presents our submission to the SoccerNet 2025 Game State Reconstruction Challenge, specifically addressing the camera calibration subtask. Our method refines homography estimation across video frames using a tracking-based approach. Reliable initialization frames are identified by comparing a U-Net pitch segmentation prediction with one derived from the NBJW-predicted homography, using their Intersection over Union (IoU) as a confidence measure. Keypoints from the NBJW model, along with auxiliary points sampled via geometric priors around randomly selected pitch locations, are projected into pitch coordinates and tracked using Lucas-Kanade optical flow. These tracked correspondences enable homography updates in subsequent frames via RANSAC. This iterative process improves temporal consistency and robustness. Evaluation using GS-HOTA and LocSim metrics shows improvements of up to 4\% over the NBJW baseline, particularly in frames lacking prominent pitch features. Future work will focus on analyzing failure cases and further fine-tuning the system to enhance performance.

\end{document}